# JutePestDetect: An Intelligent Approach for Jute Pest Identification Using Fine-Tuned Transfer Learning

[1] Md. Simul Hasan Talukder; Electrical and Electronic Engineering; Rajshahi university of Engineering and Technology, Rajshahi, Bangladesh; Email: simulhasantalukder@gmail.com

[2] Mohammad Raziuddin Chowdhury; Jahangirnagar University, Bangladesh; Email: razichy3@gmail.com

[3] Md Sakib Ullah Sourav; Shandong University of Finance and Economics, China; Email: sakibsourav@outlook.com

[4] Abdullah Al Rakin; BRAC University, Bangladesh; Email: abdullahalrakin@gmail.com

[5] .Shabbir Ahmed Shuvo; University of Rostock, Germany; Email: shuvo.shabbirahmed@gmail.com

[6] Rejwan Bin Sulaiman; Northumbria University, UK; Email: rejwan.binsulaiman@gmail.com

[7] Musarrat Saberin Nipun; Brunel University London, UK; E-mail: musarrat.nipun@brunel.ac.uk

[8] Muntarin Islam; Technical University of Denmark, Denmark; Email: islammuntarin@gmail.com

[9] .Mst Rumpa Islam; Southeast University, Bangladesh; Email: mrumpaislam@gmail.com

[10] Md Aminul Islam; Oxford Brookes University, UK; Email: talukder.rana.13@gmail.com

[11] Zubaer Haque; Brandenburg University of Technology, Germany; Email: zubaer.haque@gmail.com

Corresponding Author: Md. Simul Hasan Talukder; Rajshahi University of Engineering and Technology, Rajshahi, Bangaldesh; Email: simulhasantalukder@gmail.com

**Abstract: -** In certain Asian countries, Jute is one of the primary sources of income and Gross Domestic Product (GDP) for the agricultural sector. Like many other crops, Jute is prone to pest infestations, and its identification is typically made visually in countries like Bangladesh, India, Myanmar, and China. In addition, this method is time-consuming, challenging, and somewhat imprecise, which poses a substantial financial risk. To address this issue, the study proposes a high-performing and resilient transfer learning (TL) based JutePestDetect model to identify jute pests at the early stage. Firstly, we prepared jute pest dataset containing 17 classes and around 380 photos per pest class, which were evaluated after manual and automatic pre-processing and cleaning, such as background removal and resizing. Subsequently, five prominent pre-trained models—DenseNet201, InceptionV3, MobileNetV2, VGG19, and ResNet50—were selected from a previous study to design the JutePestDetect model. Each model was revised by replacing the classification layer with a global average pooling layer and incorporating a dropout layer for regularization. To evaluate the models' performance, various metrics such as precision, recall, F1 score, ROC curve, and confusion matrix were employed. These analyses provided additional insights for determining the efficacy of the models. Among them, the customized regularized DenseNet201-based proposed JutePestDetect model outperformed the others, achieving an impressive accuracy of 99%. As a result, our proposed method and strategy offer an enhanced approach to pest identification in the case of Jute, which can significantly benefit farmers worldwide.

## 1. Introduction

Jute is a natural fiber that has been used for thousands of years as a material to weave and make cords and is the second most important vegetable fiber after cotton [1]. India, Bangladesh, China, and Thailand are the pioneers in jute production. Around 98% of jute is produced in India and Bangladesh worldwide [2]. Asian countries like Bangladesh, Pakistan, India etc., process and prepare many goods from Jutes [2]. In Bangladesh, jute, commonly known as golden fiber, is the main cash crop. The jute industries are playing a pivotal role in our country's economy from the twentieth century till now [3]. It is grown in every district of Bangladesh, but Faridpur, Tangail, Jessore, Dhaka, Sirajganj, Bogra, and Jamalpur are the predominant jute production [3]. The livelihood of most of the farmers in these regions depends on jute production. Besides, thousands of workers are employed in the jute mills, whose only source of livelihood is jute production and the manufacture of jute goods. But there is a challenge of jute cultivation, which is the adverse effect of the pest's attack. Most often, our farmers cannot accurately diagnose what kinds of pests have attacked and, thereby, cannot take proper action. As a result, jute production is often hampered. So, the detection of pest attacks at the earliest possible stage can significantly improve yield and fiber quality. The potential crop yield loss due to pests is about 18% [4]. The financial and environmental implications of diagnosing pest attacks from a distance are numerous. Traditional methods of detecting pest attacks involve high human labour and labour costs, but they never guarantee concrete results. It is an unrealistic goal to try to detect such minor anomalies of pest attacks in their early stages on a vast land. As a result, pesticides are applied to manage agricultural losses in the later stages of pest attacks. This adds to the cost of production. Excessive use can also harm humans and other species [5], [6]. Hence, adopting better technology to detect pest attacks is imperative for sustainable agriculture and the environment. Artificial intelligence (AI) in the early identification and detection of the type's pests may solve the issue and enhance the quality of jute production. Many researchers have performed pest identification and disease diagnosis on different crops [7]. In the case of jute pest detection, previously, only four types of pest identification were performed using transfer learning [7]. But there are around 19 kinds of pests that can attack the jute crops. In order to resolve this limitation and to design a model with high accuracy and robustness of the automatic jute identification system, our team has carried out the research. But there also have many challenges in designing an AI-based jute pest identification system. First, dataset preparation is a significant issue. In our work, it is completed properly. But another is the small size of the dataset and getting robust and highly accurate output from the conventional AI system. CNNs are capable of automatically learning hierarchical features from jute images, leading to higher accuracy than traditional methods [8]. But it degrades its performance with limited data. So, we have used the transfer learning concept, which allows the user to utilize an existing trained model to develop new models that utilize fewer data samples quickly; this makes them particularly well suited to pest detection tasks with limited training data [7].

To solve the practical challenges of pest attacks, after a six-month study of around 100 and more articles, papers, blogs and hard work, we have prepared our dataset having 17 types of pest classes.

Our concern was to build a fast, robust, and reliable pest classification system based on artificial intelligence, enabling farmers to make early, informed decisions regarding the management of pests detected. This endeavor can significantly reduce the dependency on specialist feedback while also eliminating time constraints in the process. We used transfer learning to migrate a well-trained model for real-time pest classification. The contributions of our work are described in section 1.1.

### 1.1. Contribution

The main contributions of this research are as follows:

- Creation of a comprehensive jute pest dataset consisting of 17 distinct pest classes, facilitating a more thorough analysis.
- Introduction of the JutePestDetect model, designed to deliver high performance and robustness in pest identification.
- Execution and evaluation of five pre-trained models (DenseNet201, InceptionV3, MobileNetV2, VGG19, and ResNet50) on the prepared dataset, showcasing their effectiveness in jute pest detection.
- Implementation of a generalization and regularization approach, incorporating a global average pooling layer and dropout layer respectively, to enhance the learning capabilities of the transfer learning models.
- Adoption of fine-tuning techniques to fine-tune the pre-trained models specifically for jute pest identification, further improving their accuracy and adaptability.
- Thorough evaluation of the models, comparing their performance with existing works in the field, to provide a comprehensive assessment of their efficacy and effectiveness.

The remainder of this study is structured as follows. Section 2 elaborates on the work done on intelligent pest classification tasks using the transfer learning network. Details of our methodology are described in Section 3. Data collection techniques and dataset details for the dataset are listed in Section 3.1. Section 4 illustrates the results. Finally, in Section 5, conclusions are drawn from the results.

## 2. Literature Review

The use of transfer learning (TL) to enhance deep networks or use it as a feature extractor has gained rapid popularity in many domains [9]. TL has demonstrated its effectiveness in solving problems and tasks like malware classification [10], handwritten digit recognition [11], time series data classification [12], and many others.

In recent years, increasing numbers of models have been developed by researchers using different CNNs, but there are very few studies that focus on regionally significant insects identifying jute pests using AI. Sourav et al. [7] suggested a deep learning approach based on TL that can identify the four most significant Jute pest groups, namely Field Cricket, Spilosoma Obliqua, Jute Stem Weevil and Yellow Mite. Jute is a fiber from which many biodegradable products can be made that also plays a significant role in a few countries' economies, such as Bangladesh, China and India. Mallick et al. [13] developed a CNN-based deep learning model that can detect and classify six different diseases and four pests related to mung beans, a major crop in India that a substantial percentage of countrymen depend on. For the Wang, Xie, Deng, and IP102 datasets, Kasinathan et al.'s [14] pest detection method employs foreground extraction and contour recognition to find insects in a very complex background. The experiment, which produced the best detection performance with the least computation time, was conducted using shape features and machine learning techniques such as artificial neural networks (ANN), support vector machines (SVM), k-nearest neighbour (KNN), naive Bayes (NB), and CNN models. The rate of classification rate was the highest, 91.5%. Furthermore, survey research [15] examined 33 studies on automatically identifying insects from captured images and demonstrated various DL techniques, primarily using several modified CNN models to identify common insects, plants, leaves, and fruit diseases.

The goal of [16] is to benefit banana farmers by developing an AI-based banana disease and pest detection system based on a DCNN. This work developed large datasets of expert-screened banana disease and pest symptom/damage photographs gathered from various locations in Africa and Southern India. The research demonstrated that ResNet50 and InceptionV2-based models outperformed MobileNetV1. These designs reflect the most recent advances in banana disease and pest detection, with more than 90% accuracy in most of the models examined. Based on these findings, we applied the same model in our research to see how it identified jute pests. Dawei et al. [17] proposed a transfer-learning-based diagnostic system for pest detection and recognition. Pattnaik et al. [18] provide a deep CNN-based transfer learning system for categorizing pests in tomato plants. Cao et al. [19] present a transfer learning-based approach to common insect recognition to address the issue. A classification model of convolution neural network based on transfer learning technology is proposed to extract the leaf features of soybean diseases and insect pests, reduce the complexity of the image model of soybean diseases and insect pests, and solve the overfitting phenomenon caused by an insufficient number of samples. The bottom features are gradually upgraded to abstract high-level features by using the multi-layer structure of the convolution net. Excellent attribute learning capacity [20]. To improve the learning capability for pest images with cluttered backgrounds, ImageNet and MobileNet-V2 as the backbone network, and the attention mechanism, as well as a classification activation map (CAM), were incorporated in the architecture to learn the significant pest information of input images [21]. Deep convolutional neural networks (DCNN) are utilized to identify ten types of pests found in paddy crops [22]. Xin et al. [23] improve the fast-RCNN approach, which excels in target segmentation. One of the main challenges of deep learning-based algorithms is obtaining a dataset for the appropriate data class. There need to be better-structured datasets required to train CNNs for each identical insect or pest data class that is to be found around different corners of the world. However, there are several insects and pest datasets that caught much attention in recent times. The National Bureau of Agricultural Insect Resources (NBAIR) [24] is one of these, which comprises 40 classes of field crop insect photos from crops such as rice, maize, soybeans, sugarcane, and cotton. In addition, two datasets by Xie et al. in the study [25, 26] collected 24 and 40 common pests of field crops that foster pest recognition trends using CNN. A comparatively recent study that was carried out on six species of the Cicindelinae subfamily (tiger beetles) of the order Coleoptera was represented in a dataset created by Abeywardhana et al. [27] with an uneven and constrained number of pictures. Although they acknowledge that even for the trained human eye, categorizing tiger beetles can be challenging, the authors put a lot of work into taking photographs of tiger beetles from different sources, viewpoints, and scales.

The report [28] published by the Department of Agricultural Extension, Bangladesh, in June 2019 identifies a total of 19 pests that are responsible for damage to Jute production at a large scale. Along with the four Jute pest classes developed in [7], we expand the dataset to 17 classes in this work. Researchers manipulate different CNNs to implement TL with parameter modifications to achieve their proposed solution. Popular CNNs used in TL are AlexNet, LeNet, ResNet,

GoogleNet, MobileNet, and their variants. Therefore, to efficiently complete the task of classifying jute pests, we envisage employing several CNNs to apply the TL approach.

## 3. Martials and Methodology

In our work, we have proposed a robust and effective transfer learning based JutePestDetect model to detect the jute pest at the early stage of its attack on the jute plant. The architecture of the proposed methodology is shown in Figure 3. Dataset preparation, pre-processing, augmentation, designing of base model and model performance evaluation are the essential steps of this research work. All the steps are described here, step by step.

### 3.1. Dataset Preparation

One of the challenges in image classification is dataset preparation. In some fields, like medicine and agriculture, it is very difficult to collect related images. In our work, we have prepared 13 classes of jute pest images using python open-source libraries to collect images from the web. But in this case, some difficulties, such as ambient light variation, image distortion, occlusion, perspective differences, intraclass alterations, background clutter, and so on, have been raised. In our laboratory, we manually checked the quality of all the images. In the meantime, grayscale images are read in and then transformed into RGB images by using OpenCV functions in order for the image to be able to be fitted to the input layer of the proposed CNN. Different preprocessing techniques have been carried out, such as cleaning, enhancing contrast, removing background, and removing noise. The Regions of Interest (ROI) patches from each of the pest picture datasets are preprocessed to a size of 256 x 256 x3. With the help of an entomologist and one agriculturist, our team has categorized the images into different classes. The pseudocode for the preprocessing of the proposed models is included in Algorithm 1.

**Algorithm 1. Pseudo-Code of data preprocessing**

| | |
|---|---|
| 1 | **Input:** Pest data / images |
| 2 | **Output:** Preprocessed resize image |
| 3 |     **for** $x \leftarrow 1$ to n do    # $x$ = *number of images* |
| 4 |         Label= Split label from the sample and store as matrix |
| 5 |         **for** $image \leftarrow 1$ to x do |
| 6 |             Call Preprocess image function |
| 7 |             $Ri \leftarrow$ Read image |
| 8 |             $Ci \leftarrow$ Convert image in to RGB |
| 9 |             $Cj \leftarrow$ Cleaning the image |
| 10 |             $Ei \leftarrow$ Enhancing contrast |
| 11 |             $Ri \leftarrow$ Removing background |
| 12 |             $Rj \leftarrow$ Removing noise |
| 13 |             $Rs \leftarrow$ Resize image in to $256 \times 256$ |
| 14 |             $D \leftarrow$ Append image data |
| 15 |         **End** |
| 16 |         $Cd \leftarrow$ convert data into matrix |
| 17 |         $Cl \leftarrow$ Convert label into categories |
| 18 |     **End** |
| 19 | **End of Pseudo-Code** |

The labelled dataset is augmented with additional parameters using Keras ImageDataGenerator to increase the size and diversity of the dataset. The augmentation techniques are shown in Figure 1. Finally, the 13-class dataset is combined with the publicly available jute pest dataset, which has four types of pest classes. Thus, the ultimate dataset used in our work has been prepared. The total number of images in our dataset is 6209, divided into 17 classes, and the average of them in each class is around 388. The distribution of our dataset is drawn in Figure 2.

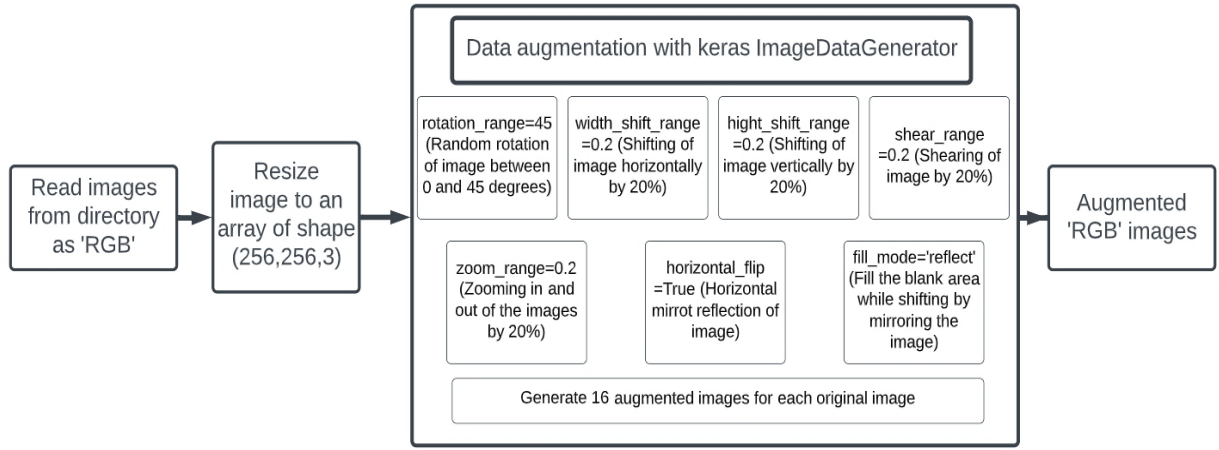

**Figure 1**. Augmentation techniques in the stage of data preparation

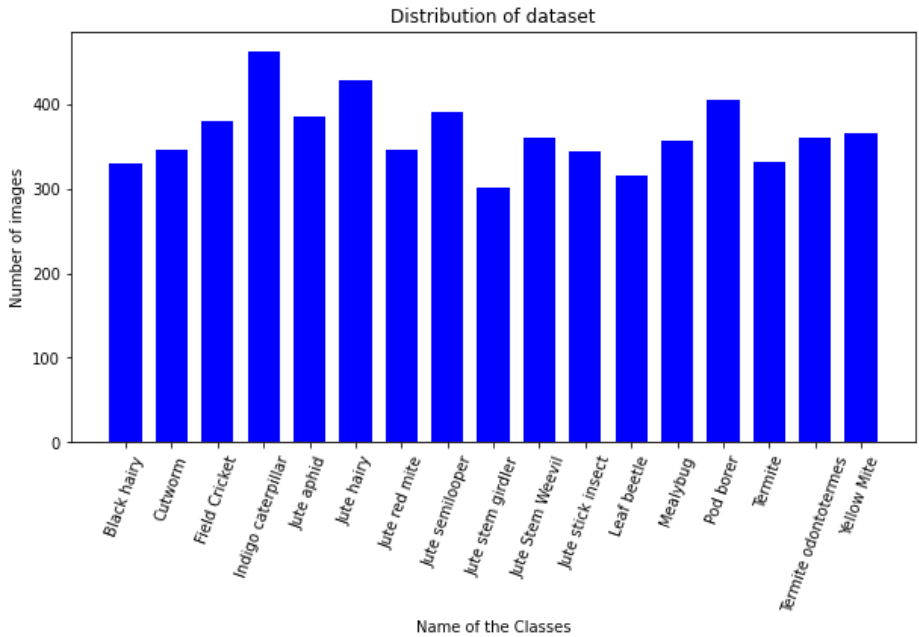

**Figure 2**. Distribution of dataset.

## 3.2. Proposed model.

In this work, we have proposed a robust jute pest identification system (JutePestDetect) that is depicted in Figure 3. After preprocessing, augmentation techniques are employed on the training and validation datasets to increase the size of the dataset and reduce model overfitting. Algorithm

2 shows the pseudocode for data augmentation techniques used in this study. For instance, for up to 20 iterations, the data sample is generated, and each iteration creates 16 new samples.

| **Algorithm 2. Pseudo-Code of Data Augmentation** | |
|---|---|
| 1  | **Input**: read original image samples x using OpenCV. |
| 2  | Resize image into 224 × 224. |
| 3  | Store resize image as an array inside a list. |
| 4  | Call Image data generator function |
| 5  | **for** $n \leftarrow 1$ to 20 **do** |
| 6  |    rotation_range = 30 |
| 7  |    width_shift_range = 0.2 |
| 8  |    height_shift_range = 0.2 |
| 9  |    zoom_range = 0.2 |
| 10 |    vertical_flip = True |
| 11 |    horizontal_flip = True |
| 12 |    Save to directory |
| 13 |    Save format as ''jpg'' |
| 14 | **End** |
| 15 | **End of Pseudo-Code** |

In our proposed system, modified versions of five prominent pre-trained CNN models, namely DenseNet201, InceptionV3, MobileNetV2, VGG19, and ResNet50, are used for feature extraction and classification. The pre-trained DenseNet201, InceptionV3, MobileNetV2, VGG19, and ResNet50 models have already been trained on the ImageNet dataset, which contains 1.2 million images in 1000 categories. The models are used to transfer the learnt weights and parameters into the jute pest dataset shown in Figure 4.

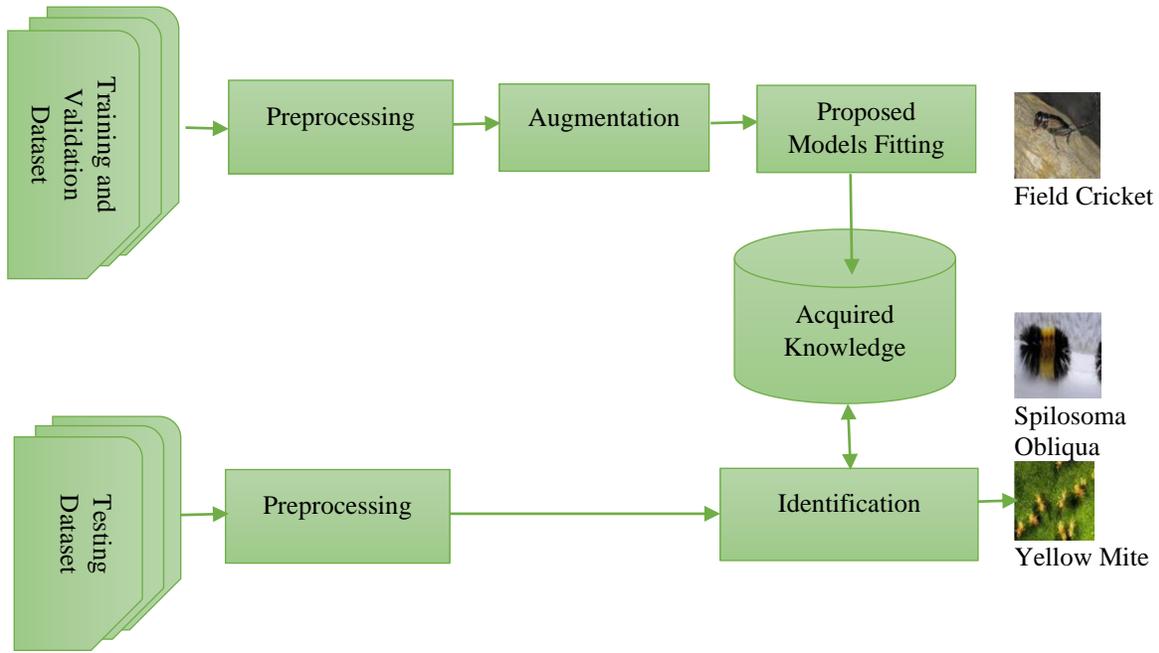

**Figure 3**. Proposed Jute Pests Identification System.

Our used models are the customized version of the five pre-trained models. We have modified the classification layer part of the models, but the base layers have remained the same, which were used for feature extraction. The classification portion includes:

- The global average pooling layer
- 30% dropout layer.
- The dense layer with SoftMax activation function showing in Equation (1) at the bottom of the pre-trained model.

The dataset is split into training, testing, and validation at the ratio of 70:15:15. Using the categorical cross-entropy loss function in equation (2), we have trained and validated the five models with training and validation data, respectively, for 50 epochs with a batch size of 16, a learning rate of 0.001, and seeds of 46. Then we evaluated each model with a test dataset, which is discussed in the result analysis section. Among the five models, modified regularized DenseNet201 has been chosen as our proposed JutePestDetect model since its performance is predominant in jute pest identification. The parameters of each modified model are listed in Table 1.

$$\sigma(\vec{z})_i = \frac{e^{z_i}}{\sum_{j=1}^{K} e^{z_j}} \qquad (1)$$

Where all the $z_i$ values are the elements of the input vector and can take any real value.

$$\text{Loss} = -\sum_{i=1}^{Output\ size} y_i . Log\ \hat{y}_i \qquad (2)$$

Where $y_i \in [1,0]$ and $\hat{y}_i$ is the actual output as a probability.

**Table 1**. Modified pretrained CNN model with the parameters.

| Model | Total Parameter | Trainable Parameters | Non-Trainable Parameters |
|---|---|---|---|
| ResNet50 | 23,622,545 | 34,833 | 23,587,712 |
| VGG19 | 20,033,105 | 8,721 | 20,024,384 |
| InceptionV3 | 21,837,617 | 34,833 | 21,802,784 |
| MobileNetV2 | 2279761 | 21777 | 2257948 |
| DenseNet201 | 18,354,641 | 32,657 | 18,321,984 |

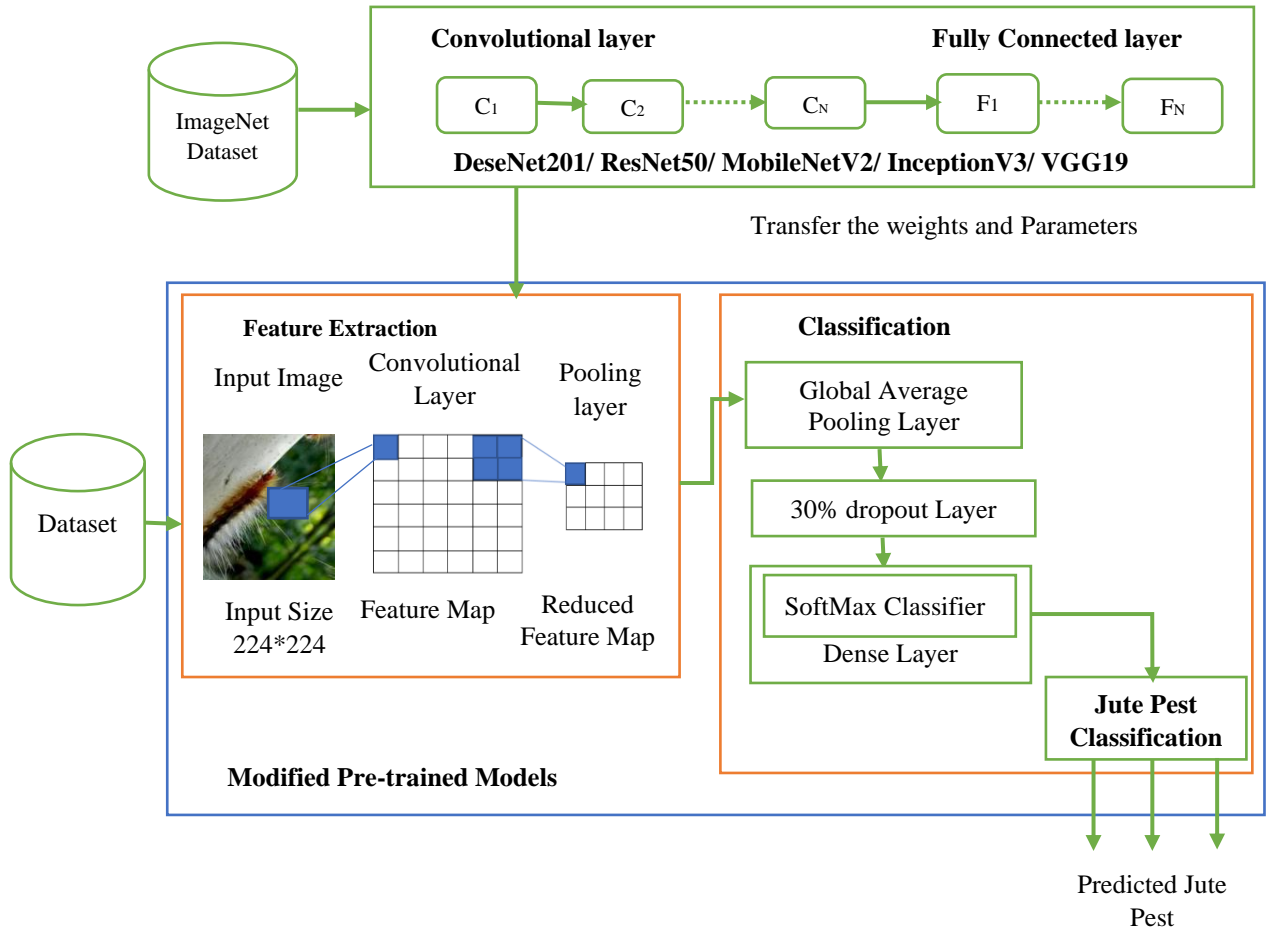

**Figure 4.** Proposed JutePestDetect Model.

### 3.3. Model Evaluation

The research work is implemented in Google Collaboratory with GPU using Keras and TensorFlow framework. In the evaluation of models, visual and quantitative analysis are the best ways. In our work, a confusion matrix is used to calculate precision, recall, F1 score, precision, and the following parameters:

(i) Accuracy: It is the ability to correctly detection of jute pests. It is given by

$$\text{Accuracy} = \frac{TP+TN}{TP+TN+FP+FN} \quad (3)$$

(ii) Precision: It represents the number of successful positive prediction models that have been created. It illustrates how much we can rely on its predictions when they are optimistic. It answers the question of how many relevant instants are returned. It is given by

$$\text{Precision} = \frac{TP}{TP+FP} \tag{4}$$

(iii) Recall: Recall is a metric that counts the number of accurate positive predictions made out of all possible positive predictions. It responds with the number of relevant instants that are retrieved. It is given by

$$\text{Recall} = \frac{TP}{TP+FN} \tag{5}$$

(iv) F1 score: It alludes to the harmonic mean of the recall and precision scores. Basically, it is utilized to compare two classifiers. The f1 score is taken into account to evaluate which delivers a desirable result when one class has higher precision, and another has stronger recall. F1 score 1 indicates that the classifier is flawless.

$$\text{F1 score} = 2 * \left(\frac{\text{Precision} * \text{Recall}}{\text{Precision} + \text{Recall}}\right) * 100 \tag{6}$$

(v) ROC AUC: ROC stands for receiving operating curve. It is the plot of true positive and false positive rate. It is a popular way to visualize model performance, and area under curve (AUC) is used to compare different models.

(vi) Macro Average: It is the arithmetic mean of each class related to precision, recall, and F1 score. It is used to evaluate overall performance of multiclass classification and can be given by

$$\text{Macro Avg Measure} = \frac{1}{N}(\text{Measure in class}_1 + \text{Measure in class}_2 + \cdots + \text{Mesure in class}_N) \tag{7}$$

Where: TP denotes the true positive; TN is the true negative; FP denotes the false positive; FN denotes the false negative. In macro-averaged, all classes equally contribute to the final averaged metric but in weighted-average, each class contribution to the average is weighted by its size.

## 4. Result Analysis and Discussion

In this portion of the research paper, the results of five different customized pre-trained models are compared and contrasted. In order to determine which of the five models provides the most accurate identification of the jute pest, the models are examined using the accuracy and loss curves, the confusion matrix, precision, recall, F1 score, accuracy, and the true and false positive rates for each of them. The whole work is performed on the Google Collaboratory platform.

The loss curve and accuracy curve of our five customized pre-trained models are depicted in Figures 5 and 6, respectively. It is found that, while analyzing the result from the accuracy and loss curve, it is observed that, for 0 to 50 epochs, the RestNet50 model does not follow a sharp increase or decrease pattern, and as a result, it provides fluctuating values both for trained and validation data. But Figures 5 and 6 have revealed that the loss curve has decreased and the accuracy curve has continuously increased with the increasing number of epochs for the remaining four models. The loss curve and the accuracy curve for both validation and training have also been optimally fitted after 50 epochs, which is in line with the basic theoretical primes. That means there has no over and underfitting. The loss and accuracy curves of the two models, InceptionV3 and VGG19, both have shown a small bit of fluctuation; however, this variation might be considered significant. Both MobileNetV2 and DenseNet201 have demonstrated the best level of consistency in the loss and accuracy curves they have presented, considering both training and validation data. Whatever the case, this research has achieved the highest training accuracy and the lowest loss for all of the four models, with the exception of RestNet50, after 50 epochs. The investigation led to the discovery of four different models: one with the best training and validation accuracy and one with the least amount of loss. From Figures 5 and 6, VGG 19 has the lowest loss at 0.56 and the highest training accuracy at 0.85. The training accuracy of all three models is comparable, coming in at 0.9730 for InceptionV3, 0.9860 for MobileNetv2 and 0.9950 for DenseNet201, respectively. DenseNet201, on the other hand, has the least amount of loss, coming in at 0.09. It is possible to deduce from the curves displayed in Figures 5 and 6 that, among the five pre-trained models, DenseNet201 is the one that is most effective in the training.

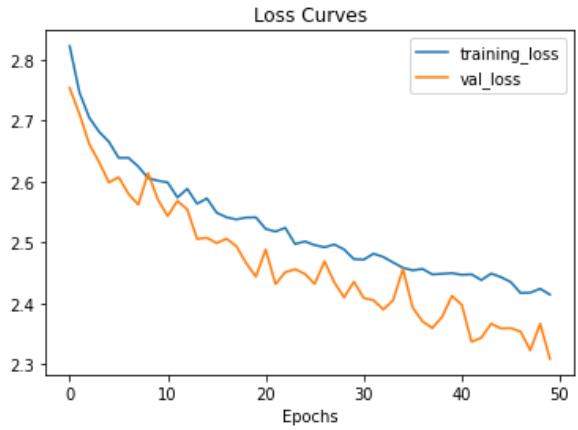
(a)

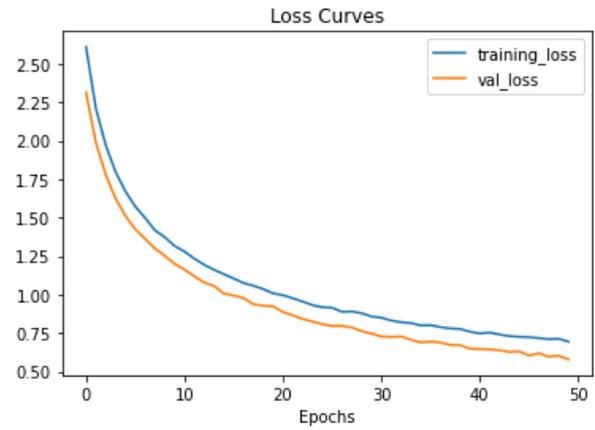
(b)

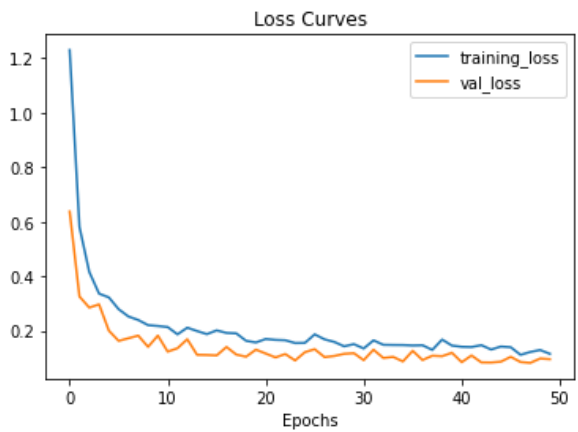
(c)

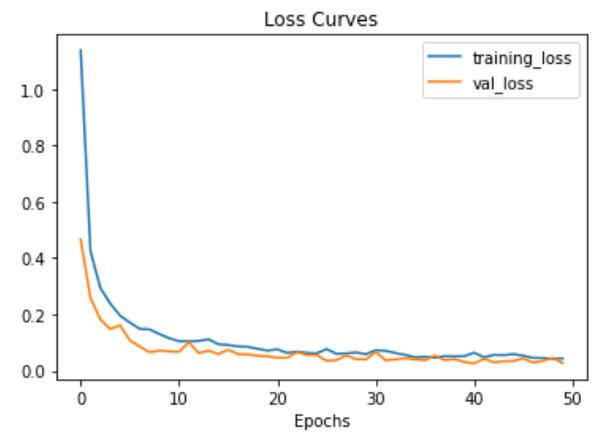
(d)

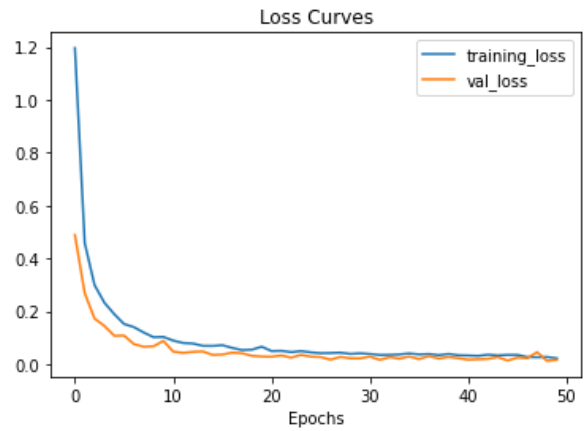
(e)

**Figure 5**. Loss curve (a) ResNet50; (b) VGG19; (c) InceptionV3; (d) MobileNetV2; (e) DenseNet201.

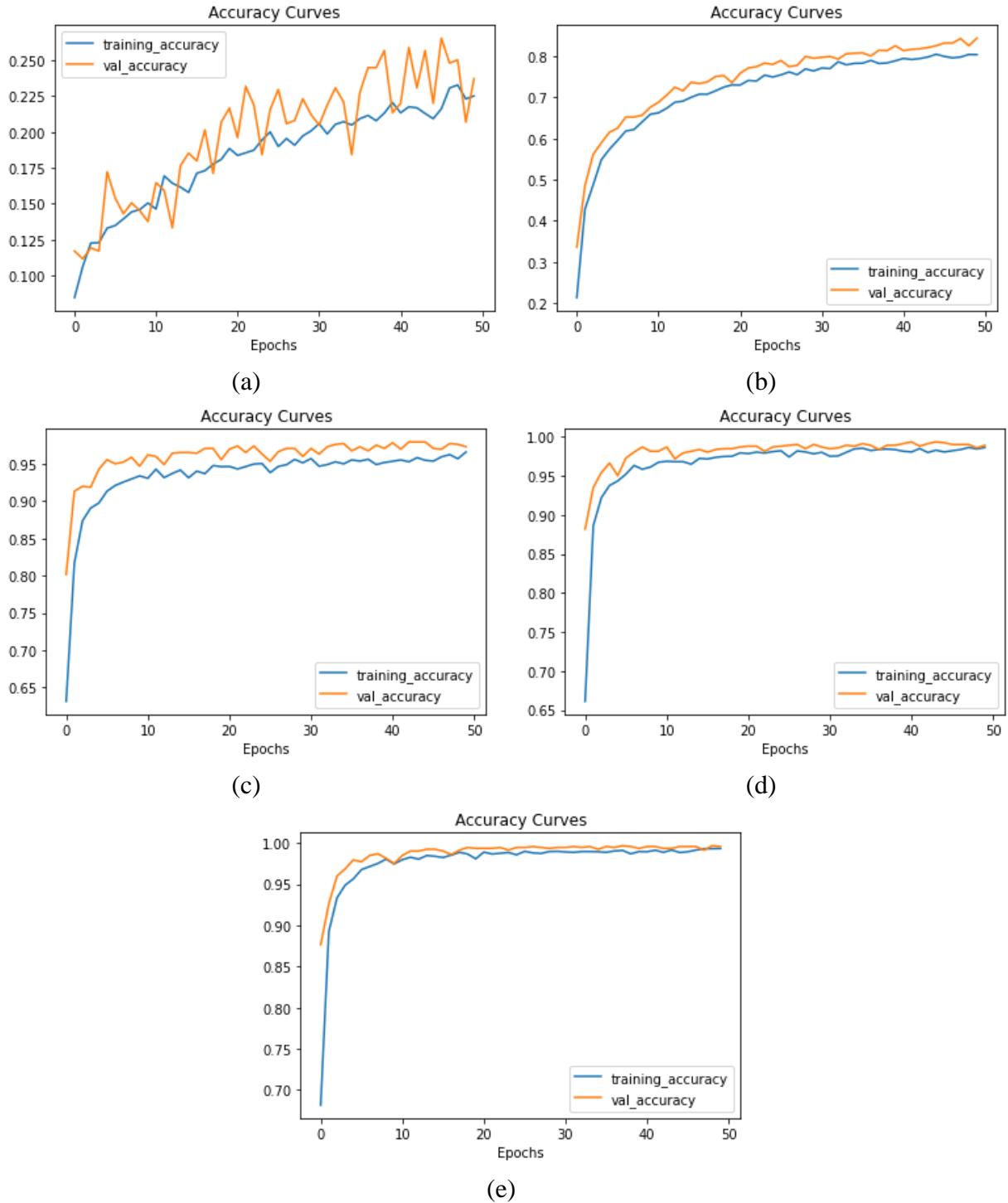

**Figure 6.** Accuracy curve (a) ResNet50; (b) VGG19; (c) InceptionV3; (d) MobileNetV2; (e) DenseNet201.

After training, all models are examined with the testing dataset in our work. Then the models are evaluated using different approaches. The confusion matrix is taken into account in our work in order to ascertain the number of misclassifications and proper classifications. The confusion matrix of each model is illustrated in Figure 7. The result of the confusion matrix displays the number of true positive, true negative, false positive and false negative of the predicted values by each model. When looking at the confusion matrix, this investigation discovered that ResNet50 has inconsistent values. The other four models have been able to classify the jute pest quite well. From Figure 7, ranking in order of classification ability, it is, of course, DenseNet201, MobileNetV2, InceptionV3, and VGG19.

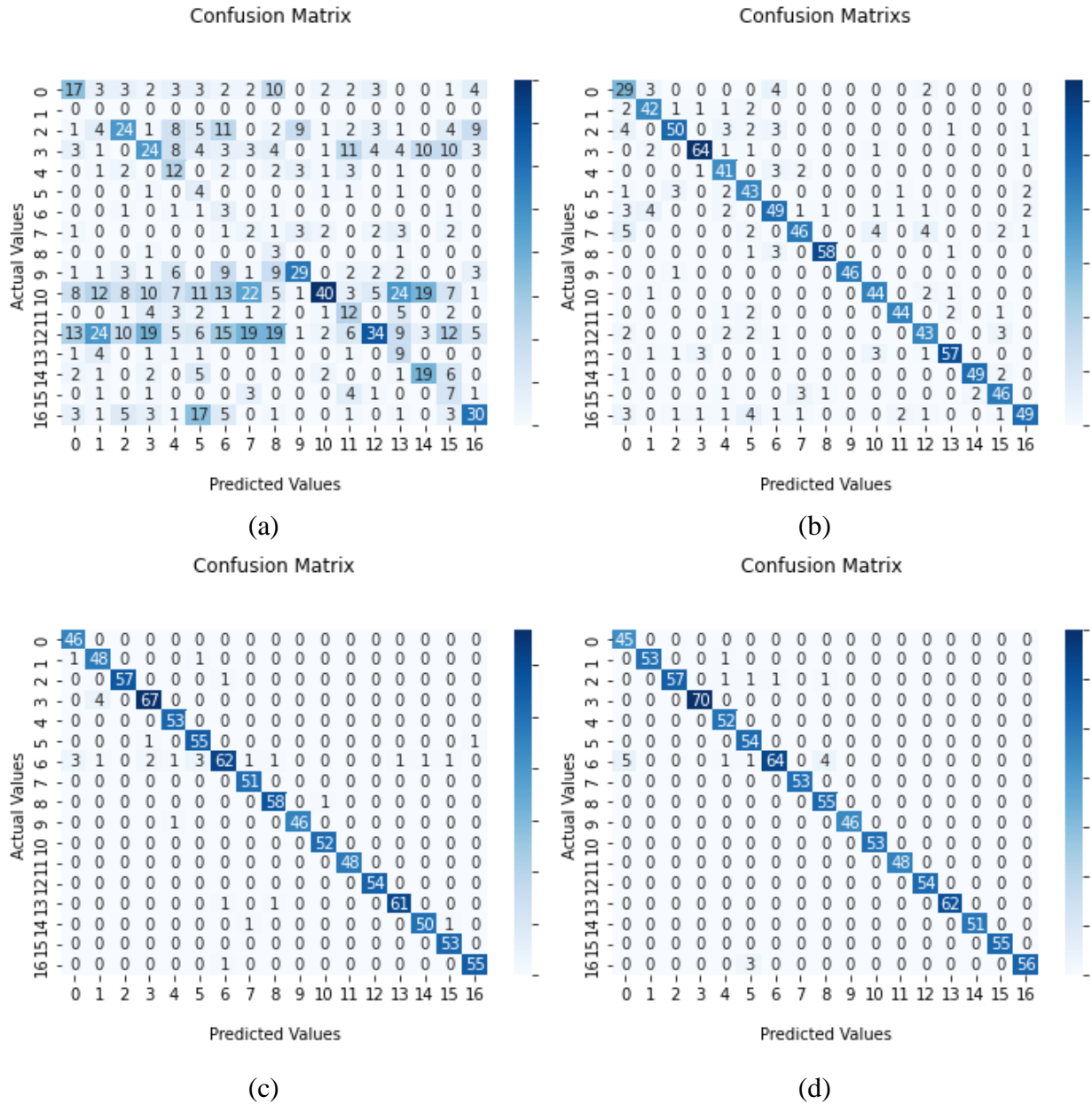

(a)

(b)

(c)

(d)

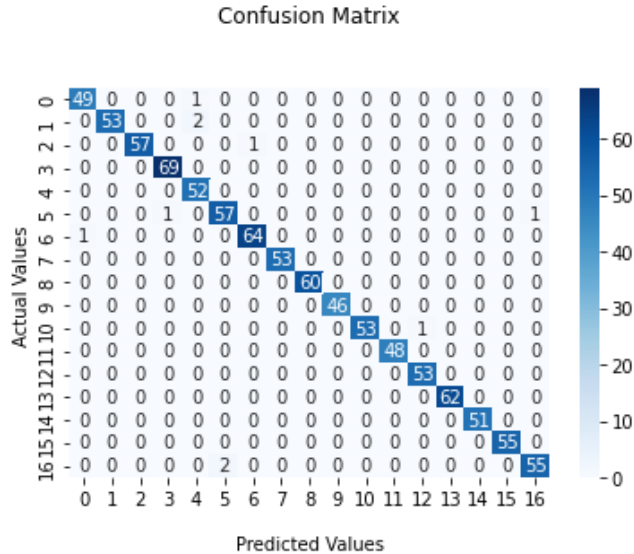

(e)

**Figure 7**. Confusion matrix (a) ResNet50; (b) VGG19; (c) InceptionV3; (d) MobileNetV2; (e) DenseNet201.

After the visual evaluation of the models in the confusion matrix, we have calculated precision, recall, F1 score, and accuracy for the quantitative evaluation. Tables 2, 3, 4, 5, and 6 represent the precision, recall, and F1 score of each model in each class, and Table 7 summarizes the overall accuracy of each model. In Table 2 – Table 6, ResNet50 has yielded very low value of precision, recall, and f1-score. However, it has exhibited zero precision, recall, and F1 score for the class named Cutworm. Thus, it did not properly classify the jute pest dataset. The remaining four models have issued moderate to high valued precision, recall and f1-score on each class. Among them, DenseNet201 is at the top level in each class. In Table 7, the macro average measures are taken into consideration. Similarly, the ResNet50 is still in the same state. VGG19 has provided 84% precision, recall, and accuracy. InceptionV3, MobileNetV2 and DenseNet201 have improved precision, recall, f1 score, and accuracy one by one, respectively. Finally, DenseNet201 has achieved 99% precision, recall, F1 score, and accuracy. InceptionV3 and MobileNetV2 have both, to some extent, contributed to the provision of correct results. The true positive rate and false positive rate of each model is also drawn in the ROC curve which are shown in Appendix 1 and Appendix 5. Here also, the DenseNet201 model has the highest possible value for its true positives, and the ROC curve for each of the 16 classes is equal to 1. Considering all of the facts presented above, this study concludes that DenseNet201 has the best accuracy, and it is the method that is most effective method for forecasting jute pests.

Table 2. Performance of the customized ResNet50 model in individual classes.

| Class Name | Precision (%) | Recall (%) | F1 score (%) |
|---|---|---|---|
| Black hairy | 34 | 30 | 32 |
| Cutworm | 0 | 0 | 0 |
| Field Cricket | 42 | 28 | 34 |
| Indigo caterpillar | 34 | 26 | 29 |
| Jute aphid | 22 | 44 | 29 |
| Jute hairy | 7 | 50 | 12 |
| Jute red mite | 5 | 38 | 8 |
| Jute semilooper | 4 | 12 | 06 |
| Jute stem girdler | 5 | 60 | 9 |
| Jute Stem Weevil | 63 | 42 | 50 |
| Jute stick insect | 75 | 20 | 32 |
| Leaf beetle | 25 | 35 | 29 |
| Mealybug | 63 | 17 | 27 |
| Pod borer | 15 | 47 | 22 |
| Termite | 37 | 50 | 43 |
| Termite odontotermes | 13 | 39 | 19 |
| Yellow Mite | 54 | 42 | 47 |

Table 3. Performance of the customized VGG19 model in individual classes.

| Class Name | Precision (%) | Recall (%) | F1 score (%) |
|---|---|---|---|
| Black hairy | 58 | 76 | 66 |
| Cutworm | 79 | 86 | 82 |
| Field Cricket | 88 | 78 | 83 |
| Indigo caterpillar | 91 | 91 | 91 |
| Jute aphid | 75 | 87 | 80 |
| Jute hairy | 73 | 83 | 77 |
| Jute red mite | 75 | 75 | 75 |
| Jute semilooper | 87 | 72 | 79 |
| Jute stem girdler | 97 | 92 | 94 |
| Jute Stem Weevil | 100 | 98 | 99 |
| Jute stick insect | 83 | 92 | 87 |
| Leaf beetle | 92 | 88 | 90 |

| Class Name | | | |
|---|---|---|---|
| Mealybug | 80 | 81 | 80 |
| Pod borer | 92 | 85 | 88 |
| Termite | 96 | 94 | 95 |
| Termite odontotermes | 84 | 87 | 85 |
| Yellow Mite | 88 | 75 | 81 |

**Table 4.** Performance of customized InceptionV3 model on individual classes.

| Class Name | Precision (%) | Recall (%) | F1 score (%) |
|---|---|---|---|
| Black hairy | 92 | 100 | 96 |
| Cutworm | 91 | 96 | 93 |
| Field Cricket | 100 | 98 | 99 |
| Indigo caterpillar | 96 | 94 | 95 |
| Jute aphid | 96 | 100 | 98 |
| Jute hairy | 93 | 96 | 95 |
| Jute red mite | 95 | 81 | 87 |
| Jute semilooper | 96 | 100 | 98 |
| Jute stem girdler | 97 | 98 | 97 |
| Jute Stem Weevil | 100 | 98 | 99 |
| Jute stick insect | 98 | 100 | 99 |
| Leaf beetle | 100 | 100 | 100 |
| Mealybug | 100 | 100 | 100 |
| Pod borer | 98 | 97 | 98 |
| Termite | 98 | 96 | 97 |
| Termite odontotermes | 96 | 100 | 98 |
| Yellow Mite | 98 | 98 | 98 |

**Table 5.** Performance of customized MobileNetV2 model on individual classes.

| Class Name | Precision (%) | Recall (%) | F1 score (%) |
|---|---|---|---|
| Black hairy | 90 | 100 | 95 |
| Cutworm | 100 | 98 | 99 |
| Field Cricket | 100 | 93 | 97 |
| Indigo caterpillar | 100 | 100 | 100 |
| Jute aphid | 95 | 100 | 97 |
| Jute hairy | 92 | 100 | 96 |
| Jute red mite | 98 | 85 | 91 |
| Jute semilooper | 100 | 100 | 100 |
| Jute stem girdler | 92 | 100 | 96 |

| Class Name | Precision (%) | Recall (%) | F1 score (%) |
|---|---|---|---|
| Jute Stem Weevil | 100 | 100 | 100 |
| Jute stick insect | 100 | 100 | 100 |
| Leaf beetle | 100 | 100 | 100 |
| Mealybug | 100 | 100 | 100 |
| Pod borer | 100 | 100 | 100 |
| Termite | 100 | 100 | 100 |
| Termite odontotermes | 100 | 100 | 100 |
| Yellow Mite | 100 | 95 | 97 |

**Table 6**. Performance of the customized DenseNet201 model on individual classes.

| Class Name | Precision (%) | Recall (%) | F1 score (%) |
|---|---|---|---|
| Black hairy | 98 | 98 | 98 |
| Cutworm | 100 | 96 | 98 |
| Field Cricket | 100 | 98 | 99 |
| Indigo caterpillar | 99 | 100 | 99 |
| Jute aphid | 95 | 100 | 97 |
| Jute hairy | 97 | 97 | 97 |
| Jute red mite | 98 | 98 | 98 |
| Jute semilooper | 100 | 100 | 100 |
| Jute stem girdler | 100 | 100 | 100 |
| Jute Stem Weevil | 100 | 100 | 100 |
| Jute stick insect | 100 | 98 | 99 |
| Leaf beetle | 100 | 100 | 100 |
| Mealybug | 98 | 100 | 99 |
| Pod borer | 100 | 100 | 100 |
| Termite | 100 | 100 | 100 |
| Termite odontotermes | 100 | 100 | 100 |
| Yellow Mite | 98 | 96 | 97 |

**Table 7**. Overall performance (Macro Avg.) of the customized pre-trained model in our dataset

| Model Name | Precision (%) | Recall (%) | F1-score (%) | Accuracy (%) |
|---|---|---|---|---|
| ResNet50 | 29 | 34 | 25 | 28 |
| VGG19 | 84 | 84 | 85 | 84 |
| InceptionV3 | 97 | 97 | 97 | 97 |
| MobileNetV2 | 98 | 98 | 98 | 98 |
| **DenseNet201** | **99** | **99** | **99** | **99** |

## 5. Conclusion and future work

Jute pest identification is a demand for farmers to ensure a good yield of marketable raw jute like other crops. To ease up the process, artificial intelligence, like the image recognition technique, might be an effective tool which has been identified by the undertaken study using DCNN and TL. The results of the experiment and comparison of different deep-meta-architectures with feature extractors showed how this deep-learning-based detector could successfully identify 17 different categories of pests. Images of 2 classes named Hooded Hopper (Otinotus Elongates Distant) and Leaf Miner (Trachys Pacifica Kerr) because of not having images by web-scrapping. Data augmentation and web scrapping has contributed to the preparation of image dataset and can lead to further analysis to build up models. Densenet201 has illustrated the highest level of accuracy at 99% over the other techniques. The acceptability of the techniques has been passed through precision, f1 score, recall, support vector, confusion matrix, loss curves, and accuracy curves.

### 5.1. Limitations

Our research team has tried its best to develop a robust and reliable AI jute pest identification system and the system has become successful. The limitations of this study are as follows:

- Exclusion of 2 types of pests and a smaller number of images.
- We were unable to undertake a professional pilot test due to a lack of permission and adequate facilities. In the future, we want to conduct the pilot test in Bangladesh, where we expect permission to be considerably simpler to get.
- At the time of writing, the availability of datasets and reference literature is limited, making it difficult to evaluate the performance of our models accurately.

However, if we can get a huge number of data, it will help us with a more convincing (and accurate) model on a large scale. In future, the limitation might be resolved by future researchers having more targeted images from the field taken either by research teams or farmers. After further study and evaluation with a larger dataset, the model can be implemented commercially to enhance agricultural growth to boost jute production and its economy through relevant modification and development.

**Authors' contributions: Talukder MSH** put in a massive effort by participating in dataset preparation and implementing the research paper along with writing sections like Methodology,

Dataset preparation, and Results Analysis and Discussion. **Chowdhury MR** was very spontaneous in contributing to the dataset collection and preparation by leading the role. **Sourav MSU** contributed to the idea formulation, dataset preparation, and literature review. **Islam M** and **Islam MR** contributed to the dataset collection, preparation, and conclusion section. **Rakin AA** contributed to writing the abstract and introduction. **Islam MA** contributed to writing the abstract and LR. **Nipun MS** contributed to writing the results section. **Shuvo SA** contributed to methodology and dataset preparation, organizing, administrating, and synchronizing the research. **Rejwan Bin Sulaiman** supervised the whole research, and he contributed by leading the entire group, formal analysis and guiding the complete writing process. **Haque Z** contributed to proofreading, enhancing quality and finalizing the manuscript.

**Funding:** This research received no external funding.

**Declaration of Competing Interest:** The authors declare that they have no conflict of interest.

**Data availability:** The dataset associated with this work is currently accessible to the public through the following URL:
https://www.kaggle.com/datasets/simulhasantalukder/jutepestindentification .

Appendix-1: ROC curve of ResNet50

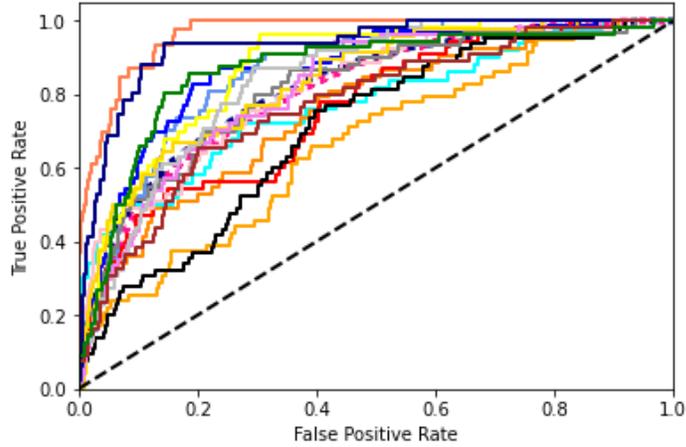

Appendix-2: ROC curve of VGG19

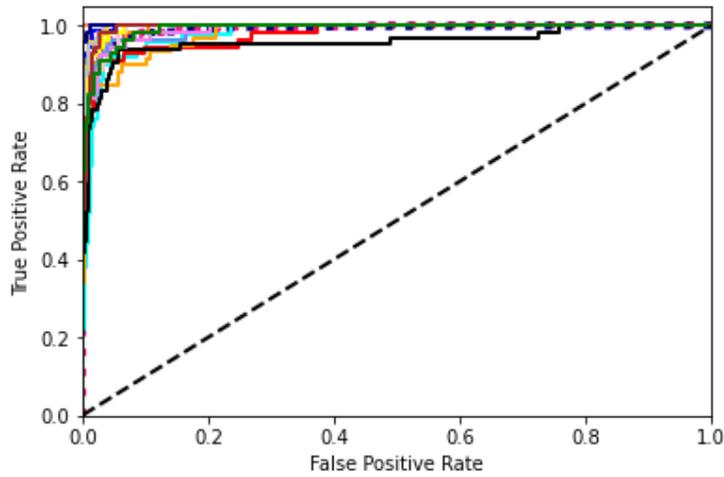

Appendix-3: ROC curve of InceptionV3

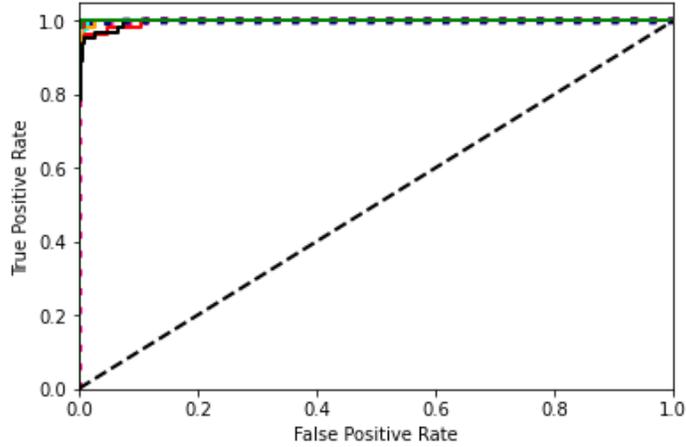

Appendix-4: ROC curve of MobileNetV2

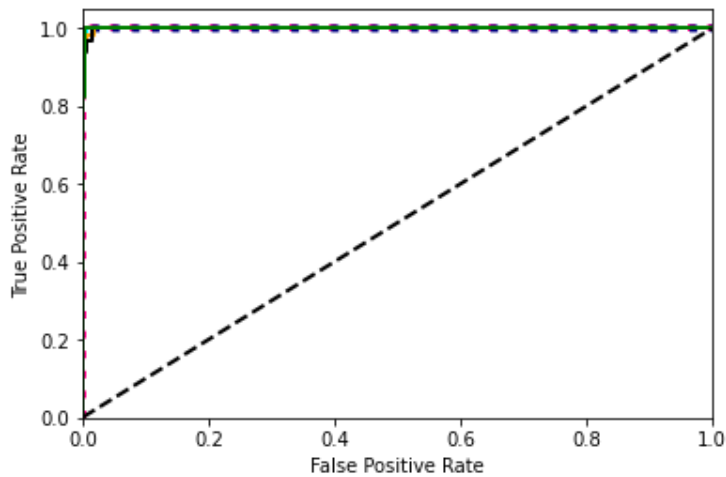

Appendix-5: ROC curve of DenseNet201

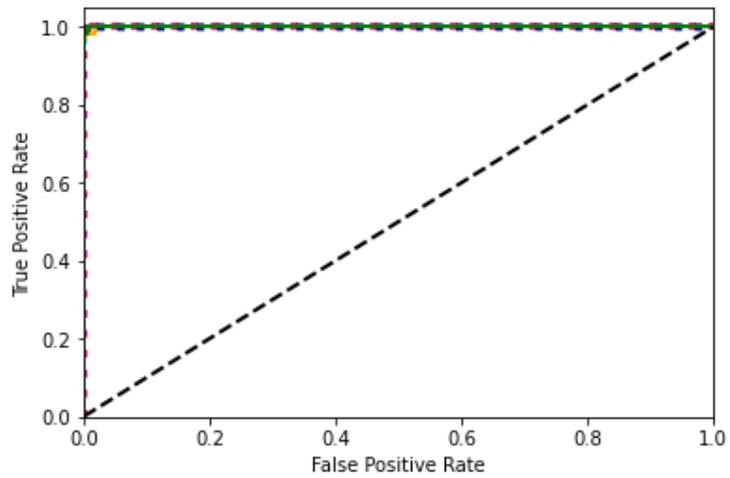